\documentclass[10pt,twocolumn,letterpaper]{article}

\usepackage{cvpr}              

\usepackage{graphicx}
\usepackage{amsmath}
\usepackage{amssymb}
\usepackage{booktabs}
\usepackage{xcolor}
\definecolor{green}{rgb}{0.1, 0.5, 0.3}
\usepackage{multirow}
\usepackage{array}
\usepackage{arydshln}
\usepackage{float}

\usepackage{listings}
\definecolor{codegreen}{rgb}{0,0.6,0}
\definecolor{codegray}{rgb}{0.5,0.5,0.5}
\definecolor{codepurple}{rgb}{0.58,0,0.82}
\definecolor{backcolour}{rgb}{0.95,0.95,0.92}
\lstdefinestyle{mystyle}{
    backgroundcolor=\color{backcolour},   
    commentstyle=\color{codegreen},
    keywordstyle=\color{magenta},
    numberstyle=\tiny\color{codegray},
    stringstyle=\color{codepurple},
    basicstyle=\ttfamily\footnotesize,
    breakatwhitespace=false,         
    breaklines=true,                 
    captionpos=b,                    
    keepspaces=true,                 
    numbers=left,                    
    numbersep=5pt,                  
    showspaces=false,                
    showstringspaces=false,
    showtabs=false,                  
    tabsize=2
}
\makeatletter
\newcommand{\thickhline}{%
    \noalign {\ifnum 0=`}\fi \hrule height 1.3pt
    \futurelet \reserved@a \@xhline
}
\newcolumntype{"}{@{\hskip\tabcolsep\vrule width 1pt\hskip\tabcolsep}}
\makeatother

\usepackage[pagebackref,breaklinks,colorlinks]{hyperref}

\usepackage[capitalize]{cleveref}
\crefname{section}{Sec.}{Secs.}
\Crefname{section}{Section}{Sections}
\Crefname{table}{Table}{Tables}
\crefname{table}{Tab.}{Tabs.}

\newcommand{\metricAbbr}{ODmAP@k\xspace}
\newcommand{\metricName}{object decorrelation\xspace}

\usepackage{amsmath,amsfonts,bm}

\def\eqref#1{equation~\ref{#1}}

\def\1{\bm{1}}

\def\vb{{\bm{b}}}

\def\vx{{\bm{x}}}
\def\vy{{\bm{y}}}

\DeclareMathAlphabet{\mathsfit}{\encodingdefault}{\sfdefault}{m}{sl}
\SetMathAlphabet{\mathsfit}{bold}{\encodingdefault}{\sfdefault}{bx}{n}

\def\gB{{\mathcal{B}}}
\def\gC{{\mathcal{C}}}
\def\gD{{\mathcal{D}}}

\def\gG{{\mathcal{G}}}

\def\gN{{\mathcal{N}}}
\def\gO{{\mathcal{O}}}

\newcommand{\mypara}[1]{\vspace{2pt}\noindent{\bf{#1}}}

\def\cvprPaperID{9} 
\def\confName{CVPR Workshop (MULA)}
\def\confYear{2023}

\begin{document}

\title{Exposing and Mitigating Spurious Correlations for Cross-Modal Retrieval}

\author{Jae Myung Kim$^{1}$, A. Sophia Koepke$^{1}$, Cordelia Schmid$^{2}$, Zeynep Akata$^{1,3}$\\ \\
$^{1}$University of Tübingen \hspace{2pt} $^{2}$Inria, Ecole normale sup\'erieure, CNRS, PSL Research University\\ $^{3}$MPI for Intelligent Systems}
\maketitle

\begin{abstract}
   Cross-modal retrieval methods are the preferred tool to search databases for the text that best matches a query image and vice versa. However, image-text retrieval models commonly learn to memorize spurious correlations in the training data, such as frequent object co-occurrence, instead of looking at the actual underlying reasons for the prediction in the image. For image-text retrieval, this manifests in retrieved sentences that mention objects that are not present in the query image. In this work, we introduce ODmAP@k, an object decorrelation metric that measures a model's robustness to spurious correlations in the training data. We use automatic image and text manipulations to control the presence of such object correlations in designated test data. Additionally, our data synthesis technique is used to tackle model biases due to spurious correlations of semantically unrelated objects in the training data. We apply our proposed pipeline, which involves the finetuning of image-text retrieval frameworks on carefully designed synthetic data, to three state-of-the-art models for image-text retrieval. This results in significant improvements for all three models, both in terms of the standard retrieval performance and in terms of our object decorrelation metric. The code is available at \href{https://github.com/ExplainableML/Spurious_CM_Retrieval}{https://github.com/ExplainableML/Spurious\_CM\_Retrieval}.
\end{abstract}

\vspace{-20pt}

\section{Introduction}
\label{sec:intro}

\begin{figure}[t]
    \centering
    \vspace{-10pt}
    \includegraphics[width=\linewidth]{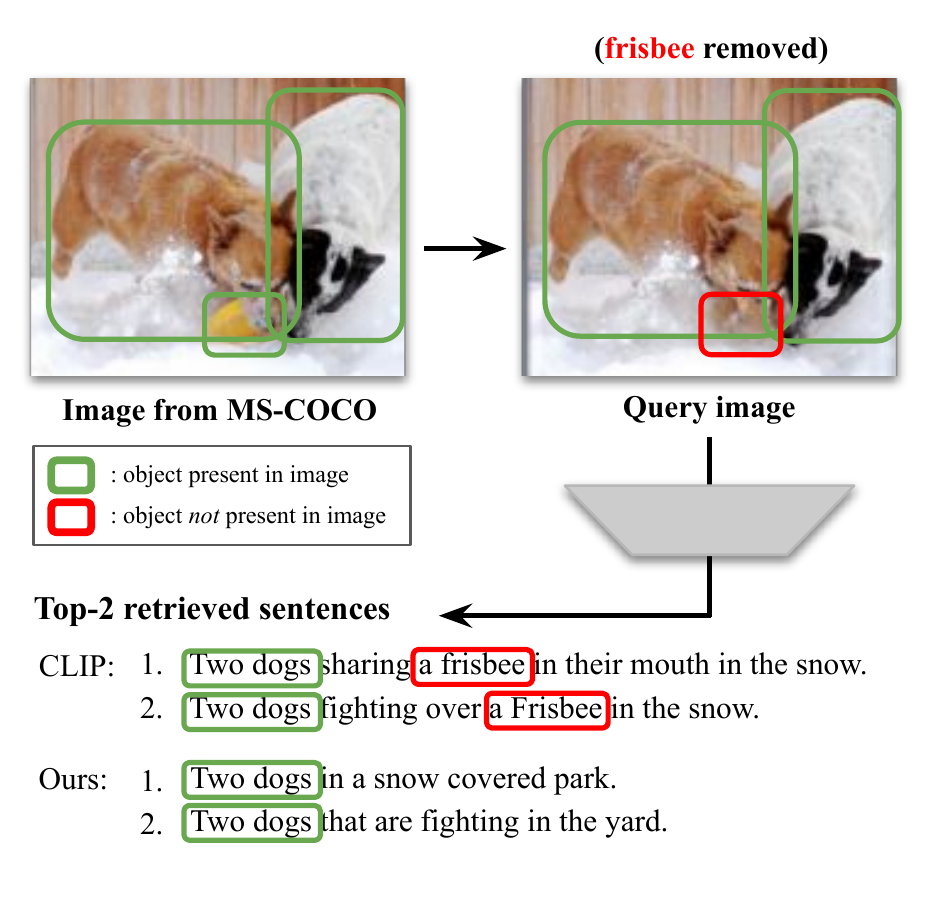} 
    \vspace{-35pt}
    \caption{\textbf{Image-to-text retrieval results with CLIP and our proposed pipeline.} 
    The query image is generated by removing an object (i.e.\ the frisbee).  
    The top-2 text retrieval results show that CLIP wrongly retrieves sentences that mention the frisbee.
    Correct and incorrect words in the retrieved sentences are marked \textcolor{green}{green} and \textcolor{red}{red} respectively. 
    \vspace{-15pt}
    \label{fig:teaser}
    }
\end{figure}

Quickly growing image and text databases necessitate an urgent need to search data efficiently. 
The cross-modal text-image retrieval task considers the setting in which query images are used to retrieve matching text from a text database and vice versa.
For instance, for a query image that shows dogs playing with a frisbee, an image-to-text retrieval model is expected to retrieve a sentence describing the scene (e.g. ``\textit{Two dogs fighting over a frisbee}'').

Commonly, image-text retrieval models are trained on paired image-text data. It has been shown that text biases and correlations in the training data translate to vision-language models (e.g.\ for image captioning~\cite{hendricks2018women}). 
In this work, we investigate the impact of spurious correlations in the training data, such as frequent co-occurrences of semantically unrelated objects, on cross-modal retrieval models trained with it.
We indeed observe spurious correlations 
in retrieved sentences describing objects not present in the query image although they tend to be somewhat related to the objects actually present in the image in the sense that they frequently appear together in the dataset.
For instance, when an image of dogs \textit{without} a frisbee is the query, the CLIP model~\cite{radford2021learning} nevertheless retrieves the sentence ``\textit{Two dogs sharing a frisbee in their mouth in the snow}'' which contains the word `frisbee' based on common co-occurrences (see Fig. \ref{fig:teaser}). 

It is easy for humans to identify failure cases where retrieved sentences mention unrelated objects. However, the standard retrieval evaluation protocols do not specifically measure these errors. To quantify those, we propose the \metricName metric \metricAbbr which captures a model's robustness to semantically unrelated object correlations in the training data. Being able to explicitly measure this specific type of error is a first step towards mitigating it since those errors also affect the retrieval task performance.

Our \metricName metric uses a designated test set with carefully designed synthetic images.
For this, object detections are used to remove commonly co-occurring objects with an inpainting framework\cite{yu2018generative}. Our \metricAbbr then measures if the retrieved text i) contains the objects that occur in the synthetic image, and ii) if the retrieved text does not mention any of the objects that have been removed from the original image and are not present in the synthetic image.
This enables us to quantify if a model has memorized commonly co-occurring objects in the training data, or if it is actually able to retrieve text that matches the objects in the query images.

Recently, \cite{agarwal2020towards} identified spurious correlations in the context of Visual Question Answering (VQA) and alleviated the biases due to spurious correlations in the training data by using synthetic training data designed for the VQA task. Similarly, we propose a finetuning pipeline that mitigates the impact of frequent co-occurrences of semantically unrelated objects in the training data on the trained models.
While data augmentation is commonly used for computer vision tasks, cross-modal augmentation is more challenging. In particular, we aim at formulating automatic augmentation strategies in both the image and text domains, whilst ensuring that the image-text pairs match and that our synthetically generated data challenges the memorization of spurious correlations in the training data. 
Our proposed finetuning pipeline improves debiasing retrieval models while having a competitive performance in a standard evaluation protocol on the MSCOCO~\cite{lin2014microsoft} and Flickr30k~\cite{young2014image} datasets.

To summarise, we make the following contributions: (1) We reveal that the performance of existing cross-modal retrieval models suffers from the existence of spurious object correlations in the training data. We propose the \metricName metric \metricAbbr to measure this correlation bias. 2) We propose a finetuning pipeline for mitigating the impact of spurious object correlations in the training data, which uses carefully designed synthesized data. 3) We demonstrate that our finetuning pipeline mitigates the model to learn the spuriousness while having a competitive performance on the standard retrieval evaluation compared with the model trained on the original dataset.

\section{Related work}
\label{sec:related_work}

\mypara{Cross-modal retrieval.}
Commonly, cross-modal retrieval methods use a learnt shared latent space to relate different modalities to one another.
This has been investigated for different modalities paired with text, such as for text-image retrieval~\cite{frome2013devise,gong2014improving,karpathy2014deep,faghri2017vse,engilberge2018finding,song2019polysemous,chen2020adaptive,thomas2020preserving,chun2021probabilistic,kiros2014unifying,karpathy2015deep,wang2016learning,eisenschtat2017linking,lee2018stacked,li2019visual,wang2019camp,zhang2020context,lu2019vilbert,li2020oscar,jia2021scaling,radford2021learning}, text-video retrieval~\cite{mithun2018learning,dong2018predicting,wray2019fine,xu2015jointly,gabeur2022masking,gabeur2020multi,nagrani2022learning,croitoru2021teachtext,bain2021frozen,aytar2008utilizing}, and text-audio retrieval~\cite{oncescu2021audio,koepke2022audio,lou2022audio,nagrani2022learning}. Furthermore, \cite{harwath2018jointly,hong2017deep,nagrani2018learnable} have explored audio-visual-text and audio-visual retrieval. In this paper, we focus on image-text retrieval.

\mypara{Image-text retrieval.}
Text-image retrieval models are usually trained to align representations across the text and image modalities for matching image-text pairs. 
Several methods have proposed different ways of quantifying the similarity between learnt visual and text embeddings~\cite{frome2013devise,gong2014improving,karpathy2014deep,faghri2017vse,engilberge2018finding,song2019polysemous,chen2020adaptive,thomas2020preserving,chun2021probabilistic}. In particular, \cite{faghri2017vse} use a triplet loss, and \cite{chun2021probabilistic} consider a probabilistic formulation.
A different line of works has developed specialized network components which allow the modeling of relations across modalities~\cite{eisenschtat2017linking,lee2018stacked,li2019visual,wang2019camp,zhang2020context}. 
Differently from the aforementioned works that aim at learning better cross-modal embeddings, we focus specifically on addressing the problem of biased cross-modal models due to spurious correlations in the training data. 

Image-text representations can be learned by using millions of image-text pairs sourced from the internet for training~\cite{radford2019language,jia2021scaling}. The contrastive alignment of images and text in the two-stream CLIP~\cite{radford2019language} and ALIGN~\cite{jia2021scaling} frameworks combined with (noisy) large-scale training data has resulted in impressive generalization capabilities. Consequently, the success of the CLIP model~\cite{radford2019language} has influenced domains far beyond text-image retrieval, as CLIP embeddings have been used for varied tasks, such as semantic segmentation~\cite{shin2022reco}, image generation~\cite{crowson2022vqgan}, and image video retrieval~\cite{luo2022clip4clip}, to name just a few.
Furthermore, several works have built on CLIP for learning strong and generalizable vision-language representations in a dual-stream fashion which allows efficient retrieval~\cite{yao2021filip,gao2022pyramidclip,alayrac2022flamingo,li2022blip,mu2022slip}. In this paper, we apply our proposed method to CLIP and to the more recent BLIP~\cite{li2022blip} which outperforms CLIP on zero-shot image-text retrieval.

\mypara{Biases in vision-language models.} 
Exposing and mitigating biases in vision-language models is of increasing research interest. Recent works consider, for instance, societal biases~\cite{berg2022prompt,wang2021gender,zhao2021understanding}, the missing correspondence of annotation~\cite{chun2022eccv_caption}, the language bias in VQA~\cite{niu2021counterfactual,chen2020counterfactual}, hubness in cross-modal retrieval~\cite{bogolin2022cross}, multimodal spurious correlations in VQA~\cite{agarwal2020towards}, the spuriousness in captioning~\cite{hendricks2018women}, object hallucination in captioning~\cite{rohrbach2018object}, large-scale vision-language pretraining~\cite{zhang2020devlbert}, or cross-modal retrieval in a unique e-commerce setting~\cite{ma2022ei}. To reduce the multimodal spurious correlations and language bias in VQA, \cite{goyal2017making,agarwal2020towards,niu2021counterfactual,chen2020counterfactual} proposed to create additional data to balance the training dataset. Inspired by the success of data augmentation in the VQA setting, we design a setup for synthesizing data that allows us to identify and mitigate biases in trained image-text retrieval models that arise from spurious correlations in the training data.

\section{Object decorrelation framework}
In this work, we propose a framework for measuring and mitigating the bias in retrieval models due to spurious object correlations in the training data. 
To examine the spuriousness of the retrieval models, we propose the \metricName metric \metricAbbr that is measured using synthetic images. In \Cref{suse:image_generation}, we describe the process of generating synthetic images. We then explain our proposed \metricAbbr metric in \Cref{suse:metric}. In \Cref{suse:pipeline}, we provide details about our finetuning pipeline for mitigating the negative impact of object correlations in the training data.

\begin{figure*}[t]
    \centering
    \includegraphics[width=\linewidth]{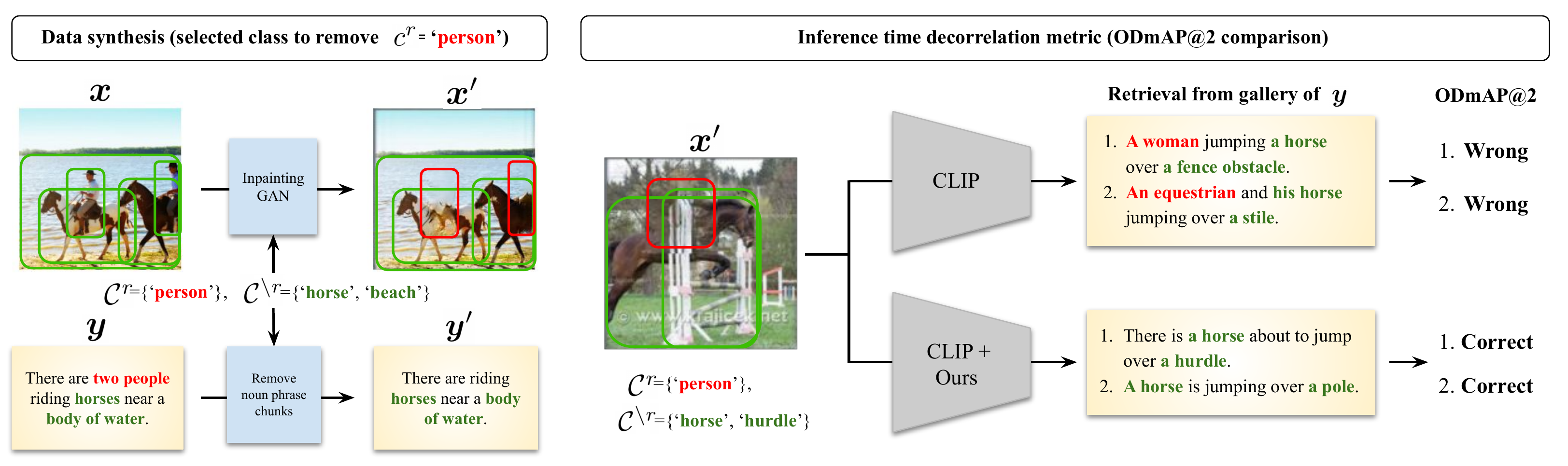} 
    \caption{\textbf{Pipeline of synthesizing the image/text pair.} The left figure shows the pipeline of how the synthetic data is generated. The right figure shows how the proposed object decorrelation metric works. Existing and nonexisting words/objects in the sentence and image are colored \textcolor{green}{green} and \textcolor{red}{red} respectively. }
    \label{fig:pipeline}
\end{figure*}

\subsection{Synthetic image generation}\label{suse:image_generation}
We consider a dataset $\gD$ consisting of image-text pairs $(\vx, \vy) \in \gD$. We examine whether the image-to-text retrieval model retrieves a sentence based on the object in a query image.
Concretely, we consider an image $\vx$ that contains several objects $\gO = \{(b_i, c_i) \,|\, i=1, \cdots, p\}$, where $b_i$ and $c_i$ refer to a box region and the class name for the $i$-th object. 
We synthesize an additional input image $\vx'$ where objects of the class $c$, $\gO_c = \{(b_i, c_i) \,|\, c_i = c \}$, are removed from the original image $\vx$. The purpose of removing the region related to class $c$ is to examine if the retrieved sentence for this query image contains the word related to class $c$. If the retrieved sentence describes the query image $\vx'$ well by not mentioning the class $c$, this model would be robust to spuriousness. The removed regions $\{b_i \,|\, c_i = c \}$ are filled in by an inpainting model~\cite{yu2018generative} to avoid the data-distribution shift which occurs when these regions are filled in by a constant value~\cite{chang2018explaining}. In the following, we describe the details of using inpainting for synthesizing images.

\mypara{Multiple object classes in the original image.} To generate a synthetic image, we select a reference image $\vx$ that contains objects from multiple classes, $|\mathrm{set}(\{c_i\}_{1}^{p})| \geq 2$. Our aim is to synthesize data that can expose and fix a model's bias towards frequently co-occurring objects. 
By removing objects of a specific class (sometimes it could be multiple classes, which we discuss below), we increase the number of images in which correlated classes do not co-occur. We propose two scenarios for synthesizing synthetic images: removing a single class and removing multiple classes.

\mypara{Removing a single class.} Assume we select a class $c^r$ which we want to remove from the image $\vx$. In this scenario, we want only the objects of class $c^r$ to be removed while others should remain in the synthesized image $\vx'$. Let us define $\gC^{\setminus r}$ as all classes in the image $\vx$ with the exception of class $c^r$, i.e.\ $\gC^{\setminus r} = \mathrm{set}(\{c_i\}_1^p) - \{c^r\}$, with the box regions of objects for the class $c^r$ and $c^{\setminus r} \in \gC^{\setminus r}$ defined as $\gB^r = \{b_i \,|\, c_i = c^r\}$ and $\gB^{\setminus r} = \{b_i \,|\, c_i = c^{\setminus r}\}$ respectively. To remove only the objects of the class $c^r$ and keep the objects of all the classes in $\gC^{\setminus r}$, the intersection between object regions should be small enough,

\begin{equation}
    \frac{ \mathrm{area}(\gB^r \cap \gB^{\setminus r}) }{ \mathrm{area}(\gB^{\setminus r}) } < \alpha_1, \;\;\;\; \text{for all} \;\; c^{\setminus r} \in \gC^{\setminus r} \, ,
\label{eq:remove_single_class}
\end{equation}
where the function $\mathrm{area}(\cdot)$ denotes the area of the given argument. If the LHS of Eq. \ref{eq:remove_single_class} is big enough, the region $\gB^r$ will overlap with a large part of the region $\gB^{\setminus r}$. If this is the case, the objects that we want to preserve in the synthetic image would also be removed when removing the region $\vb^r$. For instance, in Figure \ref{fig:pipeline} (left), we select the class `person' as the class to be removed in the reference image $\vx$. Since the `person' region is not covering the `horse' region, the inpainting model restores the removed horse region well, while the people are fully removed. In practice, we select $\alpha_1 = 0.4$.    

\mypara{Removing multiple classes.} As before, we select a class $c^r$ which we would like to remove from the image $\vx$. However, this time we want to remove objects of multiple classes, including the class $c^r$. Let us define $\gC^{r+}$ as the class set where the objects of the class $c^{r+} \in \gC^{r+}$ largely overlap with the region $\gB^r$,

\begin{equation}
    \frac{ \mathrm{area}(\gB^{r} \cap \gB^{r+}) }{ \mathrm{area}(\gB^{r+}) } > \alpha_2, \;\;\;\; \text{for all} \; \; c^{r+} \in \gC^{r+} \, ,
\label{eq:remove_multiple_class}
\end{equation}
where $\gB^{r+} = \{b_i \,|\, c_i = c^{r+}\}$. When we try to remove the region $\gB^r$ in the image $\vx$, the object information of class $c^{r+}$ would almost be lost since the region $\gB^{r+}$ highly overlaps with the region $\gB^r$. Therefore, instead of removing only the region $\gB^r$, we remove object regions of multiple classes that satisfy Eq. \ref{eq:remove_multiple_class}, i.e. $\{b_i \,|\, c_i \in \{c^r\} \cup \gC^{r+}\}$. We select $\alpha_2 = 0.8$.

\mypara{Size of the removed region.} If the removed region largely covers the image, the inpainting model would struggle to generate a realistic image. Therefore, we only consider synthetic images that satisfy:
\begin{equation}
    \frac{ \mathrm{area}(\gB') }{ \mathrm{area}(\vx) } < \alpha_3 \, ,
\label{eq:size_removed_region}
\end{equation}
where $\gB'$ is the removed region, i.e.\ $\gB' = \gB^r$ when objects of a single class are removed and $\gB' = \{b_i \,|\, c_i \in \{c^r\} \cup \gC^{r+}\}$ when objects of multiple classes are removed. We use $\alpha_3 = 0.7$.

\subsection{Object decorrelation metric \metricAbbr}\label{suse:metric}
To evaluate the bias of cross-modal retrieval models due to spurious object correlations in the training dataset, we propose the \metricName metric \textbf{\metricAbbr}. 
It measures the model's ability to capture the semantic correspondence between the query image and retrieved sentences. To measure \metricAbbr,
we use synthetic test images $\vx'$ as queries and retrieve sentences from the gallery of text captions in the dataset, $\gG_{y} = \{\vy \,|\, (\vx,\vy) \in \gD \}$. Our objective is to examine whether the retrieved sentence describes the class $c^r$ that is removed and therefore does not exist in the query image $\vx'$.

The best scenario would be that we know the matching sentence  $\vy' \in \gG_{y} $ for the synthetic image $\vx'$ since this implies that $\vy'$ would not describe the class $c^r$ that does not exist in the image $\vx'$, and we can observe whether the model retrieves the sentence $\vy'$. However, it is expensive to manually pair the synthetic image with matching sentences.
Instead, we focus on the correspondence between the object class in the image and the noun phrases in the sentence. Assume the caption $\vy$ is composed of several noun phrases $\gN = \{n_i \,|\, i=1,\ldots,q\}$. We then measure \metricAbbr by (1) checking if the sentence $\vy$ retrieved by the model does not have any of the noun phrases that is related to the class $c^r$, i.e.\ $\mathrm{match}(c^r, n_i) = 0$ for all $n_i \in \gN$, and (2) determining whether the retrieved sentence $\vy$ contains the noun phrases that are related with the class existing in the query image $\vx'$, i.e.\ $\mathrm{match}(c^{\setminus r}, n_i) = 1$ if $c^{\setminus r} \in \gC^{\setminus r} \text{ and } n_i \in \gN$, where the function $\mathrm{match}(\cdot,\cdot)$ is 1 when two arguments are related and 0 if not.   
We assume the retrieved sentence is correct if both conditions (1) and (2) are satisfied, and then measure the accuracy using the mean average precision at k (mAP@k)~\cite{musgrave2020metric}.
For instance, in Figure \ref{fig:pipeline} (right) the query image does not contain the class `person' but contains the classes `horse' and `hurdle'. If the retrieved sentence contains a noun phrase that is related to the class `person', e.g.\ `A woman' or `An equestrian', we regard this retrieved sentence as wrong. 
To reduce the possibility of no relevant and correct sentences in the gallery that describe the query synthetic image $\vx'$ well, we form the gallery $\gG$ with sentences not only from the test dataset but also from the training and validation datasets. Also, the design of the matching function $\mathrm{match}(\cdot,\cdot)$ differs on the dataset which is described in \Cref{sec:exp:setup}. 


\subsection{Finetuning pipeline}\label{suse:pipeline}
\label{sec:finetune}

In this section, we introduce a data augmentation method $\gD'$ that helps mitigate the spuriousness of the dataset and leads to debiasing the model when $\gD'$ is used for fine-tuning. We use the synthetic image $\vx'$ generated by the method introduced in \Cref{suse:image_generation} using the training dataset. To form the synthetic dataset $(\vx', \vy') \in \gD'$, we should make a caption $\vy'$ that describes the synthetic image $\vx'$. One solution would be to manually describe each image, which requires huge annotation costs. Instead, we propose three methods that automatically generate sentence with the pre-existing resources: prompts, a pre-trained captioning model, and noun phrase chunking.

\mypara{Prompts.} Language prompts have recently gained attention for boosting the performance of large language and vision-language models \cite{radford2021learning,CUPL,DCLIP}. For instance, CLIP~\cite{radford2021learning} achieved an impressive zero-shot classification performance on ImageNet~\cite{deng2009imagenet}. To predict the class label in CLIP, a prompt, e.g. ``\textit{A photo of [classname]}'', passes through the CLIP text encoder to form a text embedding and measure the distance with an image embedding. In our case, we use the prompt for describing the synthetic image $\vx'$. Assume $\gC^{\setminus r}$ is the set of classes for which objects are present in the synthetic image $\vx'$. We synthesize the caption $\vy'$ with the prompt that includes classes in $\gC^{\setminus r}$ , e.g.\ $\vy' = \text{``\textit{A photo of person and dog}''}$ when $\gC^{\setminus r} = \{\text{`\textit{person}', `\textit{dog}'}\}$. In practice we use 80 prompts and randomly select a single prompt to generate the caption $\vy'$.

\mypara{Pre-trained captioning model.} Another way to easily generate a caption for a synthetic image is to use a pre-trained captioning model. We use the ClipCap~\cite{mokady2021clipcap} image captioning model trained on the MSCOCO dataset \cite{lin2014microsoft} to generate the caption $\vy'$.

\mypara{Removing noun phrase chunks.} We can synthesize the caption $\vy'$ by removing  relevant noun phrases from the original caption $\vy$. We assume that the caption $\vy$ is composed of several noun phrases $\gN = \{n_i \,|\, i=1, \cdots, q\}$, and the synthetic image $\vx'$ is obtained by removing the objects from classes in $\gC^r$ from the original image $\vx$. Similar to how the ODmAP@k is measured, we select the noun phrases, $\gN^r \subset \gN$, that are related to the class in $\gC^r$, i.e.\ $\mathrm{match}(c^{\setminus r}, n_i) = 1$ for $ c^{\setminus r} \in \gC^{\setminus r}$ and $ n_i \in \gN^r$, and remove the selected noun phrases $\gN^r$ from the caption $\vy$ to generate $\vy'$. For instance, the noun phrases for $\vy =$ ``\textit{Two dogs fighting over a frisbee}'' are $\gN = \{\text{`\textit{Two dogs}', `\textit{a frisbee}'}\}$. Removing the class `\textit{frisbee}' gives the synthetic caption $\vy' = $ ``\textit{Two dogs fighting over}''.

Even though this process of synthesizing the sentence is similar to the method used in ODmAP@k in the sense that they both use the noun phrase that matches the class, we argue that it is not designed to explicitly improve the ODmAP@k. In the process of generating $\vy'$, the noun phrase is erased from the original sentence $\vy$, and therefore $\vy'$ might be grammatically incorrect. Instead, ODmAP@k is basically retrieving the sentence from the gallery formed with the original sentence $\vy$, and the noun phrase is used only to check whether the retrieved sentence describes a specific class when measuring the spuriousness.

\mypara{Training with the synthetic dataset.}
With the use of both datasets $\gD$ and $\gD'$, the training dataset becomes more balanced in terms of the co-occurrence of visual objects and semantics in the image/text pair. This balanced dataset can be used to mitigate the bias that arises from memorizing object co-occurrence. 

\section{Experiment}
\label{sec:experiment}

{
\setlength{\tabcolsep}{2.pt}
\renewcommand{\arraystretch}{1.2}
\begin{table*}[t]
\centering
\small
\begin{tabular}{lccccccccc}
\thickhline
& \;\; & \multicolumn{3}{c}{MS-COCO (5K test set)} & \;\;\;\; & \multicolumn{3}{c}{Flickr30k (1K test set)} \\
& & ODmAP@1 & ODmAP@5 & ODmAP@10 & & ODmAP@1 & ODmAP@5 & ODmAP@10  \\ \hline
\\[-3ex]
VSE++ \cite{faghri2017vse} & & 
56.0 & 49.3 & 45.2 & & 59.7 & 53.2 & 49.3  \\
VSE++ \cite{faghri2017vse} + Ours  & & 
\bf 59.8 & \bf 52.8 & \bf 48.3 & & \bf 62.5 & \bf 54.6 & \bf 51.5 \\ \hdashline
\\[-3ex]
CLIP \cite{radford2021learning} (zeroshot) & & 
58.6 & 51.6 & 47.1 & & 59.0 & 52.0 & 48.4 \\
CLIP \cite{radford2021learning} (finetune) & & 
59.8 & 53.2 & 48.8 & & 60.2 & 52.2 & \bf 48.4 \\
CLIP \cite{radford2021learning} + Ours & & 
\bf 70.1 & \bf 62.3 & \bf 56.8 & & \bf 61.4 & \bf 52.3 & 47.6 \\ 
\hdashline
BLIP \cite{li2022blip} (zeroshot) & & 
60.2 & 52.3 & 47.4 & & \bf 62.9 & 54.9 & 51.2 \\
BLIP \cite{li2022blip} (finetune) & & 
65.3 & 58.3 & 53.8 & & 58.3 & 52.5 & 50.3 \\
BLIP \cite{li2022blip} + Ours & & 
\bf 71.6 & \bf 63.7 & \bf 58.5 & & 62.5 & \bf 55.4 & \bf 52.4 \\ %

\thickhline
\end{tabular}
\caption{\textbf{Evaluation of object decorrelation of cross-modal retrieval on the MS-COCO (left) and Flickr30k (right) datasets.} We evaluate three different models that use different architectures (CNN, RNN, or Transformer-based) and loss functions for training the model (triplet loss, contrastive loss, or matching loss). 
}
\vspace{5pt}
\label{tbl:dbmap_evaluation}
\end{table*}
}

{
\setlength{\tabcolsep}{2.pt}
\renewcommand{\arraystretch}{1.2}
\begin{table*}[t]
\centering
\small
\begin{tabular}{lcccccccccccccccc}
\thickhline
& \;\; & \multicolumn{7}{c}{MS-COCO (5K test set)} & \;\;\;\; & \multicolumn{7}{c}{Flickr30k (1K test set)} \\
& & \multicolumn{3}{c}{image $\rightarrow$ text} & \;\; & \multicolumn{3}{c}{text $\rightarrow$ image} & & 
\multicolumn{3}{c}{image $\rightarrow$ text} & \;\; & \multicolumn{3}{c}{text $\rightarrow$ image} \\
& & R@1 & R@5 & R@10 & & R@1 & R@5 & R@10 & & R@1 & R@5 & R@10 & & R@1 & R@5 & R@10 \\ \hline
\\[-3ex]
VSE++ \cite{faghri2017vse} & & 
32.8 & 62.0 & 74.6 & & 24.1 & 52.9 & 66.3 & & 
40.9 & 68.7 & \bf 78.5 & & \bf 31.8 & 59.9 & 70.9  \\
VSE++ \cite{faghri2017vse} + Ours  & & 
\bf 35.0 & \bf 64.3 & \bf 76.2 & & \bf 24.7 & \bf 53.5 & \bf 66.6 & & 
\bf 42.5 & \bf 69.5 & 77.7 & & \bf 31.8 & \bf 60.7 & \bf 71.8 \\ \hdashline
\\[-3ex]
CLIP \cite{radford2021learning} (zeroshot) & & 
50.6 & 75.1 & 83.6 & & 30.1 & 55.7 & 66.8 & & 
79.0 & 94.3 & 98.2 & & 58.0 & 82.9 & 89.9 \\
CLIP \cite{radford2021learning} (finetune) & & 
65.5 & \bf 87.4 & \bf 93.3 & & \bf 48.6 & \bf 75.7 & \bf 84.7 & & 
84.2 & 96.1 & 98.1 & & 68.6 & 90.4 & 95.1 \\
CLIP \cite{radford2021learning} + Ours & & 
\bf 65.6 & 87.2 & 93.1 & & 48.4 & \bf 75.7 & 84.4 & & 
\bf 85.0 & \bf 96.5 & \bf 99.0 & & \bf 69.9 & \bf 90.9 & \bf 95.4 \\ 
\hdashline
BLIP \cite{li2022blip} (zeroshot) & & 
71.2 & 90.1 & 94.6 & & 55.0 & 79.3 & 86.9 & & 
85.5 & 97.9 & 99.0 & & 77.7 & 94.2 & 96.9 \\
BLIP \cite{li2022blip} (finetune) & & 
78.0 & 93.8 & 97.0 & & 61.0 & 84.2 & \bf 90.8 & & 
96.1 & \bf 99.8 & 99.9 & & 85.8 & 97.2 & \bf 98.8 \\
BLIP \cite{li2022blip} + Ours & & 
\bf 78.7 & \bf 94.5 & \bf 97.6 & & \bf 61.3 & \bf 84.4 & \bf 90.8 & & 
\bf 96.2 & 99.6 & \bf 100.0 & & \bf 86.2 & \bf 97.5 & \bf 98.8 \\ %
\thickhline
\end{tabular}
\caption{\textbf{Evaluation of standard cross-modal retrieval on the MS-COCO (left) and Flickr30k (right) datasets.} We evaluate three different models that use different architectures (CNN, RNN, or Transformer-based) and loss functions for training the model (triplet loss, contrastive loss, or matching loss). 
}
\label{tbl:standard_evaluation}
\end{table*}
}

We describe our experimental setup in \Cref{sec:exp:setup}. We then show the evaluation results on the spuriousness in \Cref{sec:exp:debiasedness} and the standard cross-modal retrieval in \Cref{sec:exp:standard}. Finally, we provide qualitative results and an analysis of our method in \Cref{sec:exp:analysis}.

\subsection{Experimental setup}
\label{sec:exp:setup}

\mypara{Cross-modal retrieval datasets.} We conduct experiments on two datasets: MS-COCO~\cite{lin2014microsoft} and Flickr30k~\cite{young2014image}. {\bf MS-COCO} contains 123,287 images and each image is manually annotated with 5 sentences. Following existing works on this benchmark, we adopted the standard evaluation split to test the general cross-modal retrieval performance, using 113,287 images for training, 5,000 images for validation and 5,000 images for testing. 
{\bf Flickr30k} contains 31,783 images from the Flickr30k website and each image is annotated with 5 sentences. We used the standard evaluation split that contains 29,000 images for training, 1,000 images for validation and 1,000 images for testing. 

\mypara{Compared methods.}
We adopt three cross-modal retrieval methods for comparison: VSE++~\cite{faghri2017vse}, CLIP~\cite{radford2021learning}, and BLIP~\cite{li2022blip}. The methods we compare to are selected to cover different architectures (CNN, RNN, or Transformer-based) and loss functions (triplet loss, contrastive loss, or matching loss). Specifically, we consider the VGG19~\cite{simonyan2014very} and Bi-GRU~\cite{cho2014properties} architectures for the image encoder and the text encoder in VSE++, and ViT-B~\cite{dosovitskiy2020image}, and BERT~\cite{devlin2018bert} in both CLIP and BLIP. Following the original papers, VSE++ is trained with a triplet loss, CLIP with a contrastive loss, and BLIP with a contrastive loss and a matching loss. We refer to the zero-shot application of the CLIP and BLIP models as CLIP (zeroshot) and BLIP (zeroshot), respectively, and to the model that was fine-tuned on the original dataset $\gD$ as CLIP (finetune) and BLIP (finetune).

\mypara{Implementation details.} We finetune the compared methods with both the original and synthetic datasets, $\gD \cup \gD'$, on a single Quadro RTX 6000 GPU. To finetune VSE++ and CLIP, we run 10 epochs with a batch size of 256. We use the Adam~\cite{kingma2014adam} optimizer with a learning rate in 1e-4 for VSE and 2e-6 for CLIP, and decay the learning rate linearly with a rate of 0.5 every 2 epochs. To finetune the BLIP, we run 2 epochs with a batch size of 16. We use AdamW~\cite{loshchilov2018decoupled} optimizer with a learning rate of 1e-5, and decay the learning rate linearly with a rate of 0.5 every half epoch. If not mentioned otherwise, we use the method of removing noun phrase chunks to synthesize the text dataset as default. When synthesizing the text by removing noun phrase chunks, we match the noun phrase with the class name of the to be removed class. The mechanism for matching these depends on the dataset. On the Flickr30k dataset, the (class of object in image, noun phrase in text) pair is annotated. On the MS-COCO dataset, we manually create a list of words that are related to the class, and regard the noun phrase and the class as a matching pair when the noun phrase contains a word that is related to the class. The manual list is provided in the Appendix. The number of classes in MS-COCO and Flickr30k is 80 and 4, respectively.

\mypara{Synthetic dataset.} With the data augmentation pipeline described in \Cref{suse:pipeline}, we synthesized 45,467 image/text pairs for the MS-COCO training dataset and 4,650 pairs for the Flickr30k training dataset. These are obtained by considering a single caption among 5 possible captions for each image in the original dataset $\gD$ to generate a synthetic image/text pair. Using all 5 (which results in about 5 times more synthetic image/text pairs) did not yield a noticeable performance improvement (discussed in \Cref{sec:exp:analysis}). Therefore, we use 45,467 and 4,650 synthetic pairs for finetuning the models on MS-COCO and Flickr30k respectively.

\begin{figure*}[t]
    \centering
    \includegraphics[width=\linewidth]{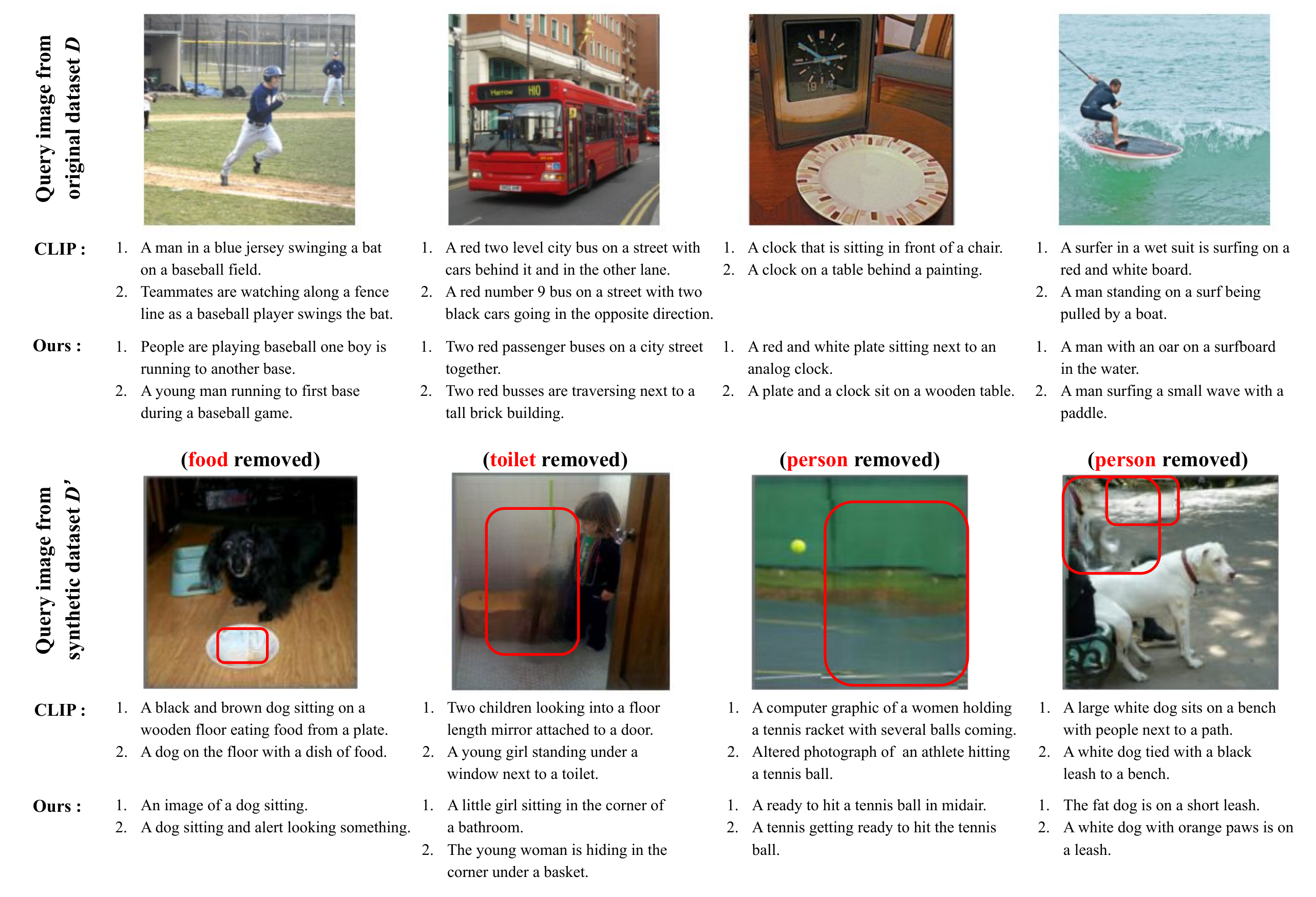} 
    \caption{\textbf{Qualitative results of image-to-text retrieval.} The first (resp. second) row shows the image-to-text retrieval results when the query image is from the original dataset (resp. synthetic dataset). For the synthetic dataset, we indicate which class has been removed from the original dataset in \textcolor{red}{red}.}
    \label{fig:qualitative}
\end{figure*}

\subsection{Evaluating object decorrelation}
\label{sec:exp:debiasedness}

We evaluate the object decorrelation of our method and four compared methods using the ODmAP@K metric on the MS-COCO and Flickr30k datasets in Table \ref{tbl:dbmap_evaluation}. We observe that our method outperforms the other frameworks that we compare to. For instance, CLIP (finetune) gives 59.8\% and 60.2\% scores in ODmAP@1 on the MS-COCO and Flickr30k datasets, respectively, while our method yields 70.1\% and 61.4\%,  outperforming CLIP (finetune) by a margin of 10.3\% and 1.2\%. Similar trends are observed across different datasets and compared methods except for BLIP on Flickr30k where BLIP (zeroshot) is 0.4\% better than ours. These results imply that our method has the ability to get better retrieval results than the compared methods based on the correct object cue present in the image. Also, we observe that the baseline model tends to have a better ability to debias the spurious correlation when its standard retrieval ability is better. For instance on the MS-COCO dataset, BLIP (finetune) has the best score on ODmAP@1 at 65.3\%, followed by CLIP (finetune) with 59.8\% and VSE++ at 56\%. Finally, we observe that for the large-scale models the fine-tuned model gives a better score than the zero-shot model. The ODmAP@1 score improves from 60.2\% to 65.3\% for BLIP and from 58.6\% to 59.8\% for CLIP when the zero-shot model is fine-tuned only on the original dataset.   

\subsection{Evaluating standard cross-modal retrieval}
\label{sec:exp:standard}

\mypara{Standard metric.} We evaluate the cross-modal retrieval performance using the recall at K (\textbf{R@K}) which measures the fraction of queries for which at least one correct sample is in the top K retrieved items. 

\mypara{Quantitative results for the standard retrieval evaluation.} We evaluate the cross-modal retrieval results using the standard retrieval metric. The results are reported in Table \ref{tbl:standard_evaluation}. Our method shows competitive performance across different datasets and baseline models. For instance, CLIP (finetune) yields 65.5\% and 48.6\% for image-to-text and text-to-image R@1 respectively on the MS-COCO dataset, while our method gives 65.6\% and 48.4\%. In both cases, the difference is less than 0.5\%. Similar trends are observed across different datasets and compared methods, where our method shows competitive or marginally better results. Overall, these results suggest that our method addressed the co-occurrence bias in the model without harming the overall retrieval performance.

\subsection{Analysis}
\label{sec:exp:analysis}

\mypara{Qualitative result.} Figure \ref{fig:qualitative} shows a qualitative comparison between CLIP \cite{radford2021learning} and CLIP fine-tuned on our augmented dataset. The first row shows the top-2 retrieved sentences given the original image as a query. It can be observed that the retrieved sentence describes an object that is not present in the query image. For instance, the first retrieved sentence by CLIP for the first image as a query contains the word `bat` which is not visible in the image but is related to other objects like the man with a blue jersey or a baseball field. This can be observed more clearly in the second row of Figure \ref{fig:qualitative} where the synthetic image is given as a query. While objects of the selected class (or classes) are removed from the original image to generate the synthetic image, we observe that CLIP still retrieves the sentence that describes the object that is removed. For instance, the top-1 retrieved sentence by CLIP for the first image as a query contains the word `food' which is the class that has been removed from the original image. Our method retrieves the sentence that describes the visible objects without mentioning other objects that are not present in the query image. 

\mypara{Impact of different methods for synthetically generating data.} Here, we explore the impact of different methods for synthetically generating text and images in $\gD'$. For the image synthesis, we consider three different methods that are commonly used for removing the information from the image: zero padding, mean padding, and blur padding. Zero padding refers to the method where the removed region $\gB^r$ in the original image is filled in with zeroes. Mean padding and blur padding refer to the compared methods when $\gB^r$ is filled in with the average pixel values of $\gB^r$ and the Gaussian blur of $\gB^r$, respectively. For text synthesis, we consider two additional methods: prompts and the captioning model to generate the text (as described in \Cref{suse:pipeline}). 

The results for different data synthesis methods are shown in Table \ref{tbl:diff_synth}. We observe that except for using prompts to synthesize text and inpainting for image synthesis, using the synthetic dataset generated by noun phrase removal and inpainting method appear to be suited best for debiasing the model.
When the image is generated using blur-, zero-, or average padding, the synthesized image would be considered out-of-distribution which would result in the model learning other biases. 
Furthermore, using prompts for text synthesis shows the best result on ODmAP@1 but drastically decreases on ODmAP@5 and ODmAP@10. We discovered that this model suffers from the hubness issue, i.e.\ the retrieved sentence tends to be the same for many different queries. We conjecture that this happens because the prompts use a similar text format, and the model learns to match this specific text format to the synthetic images. These results suggest that the careful design of data augmentation is needed to debias the model. 

\begin{figure}[t]
    \centering
    \includegraphics[width=\linewidth]{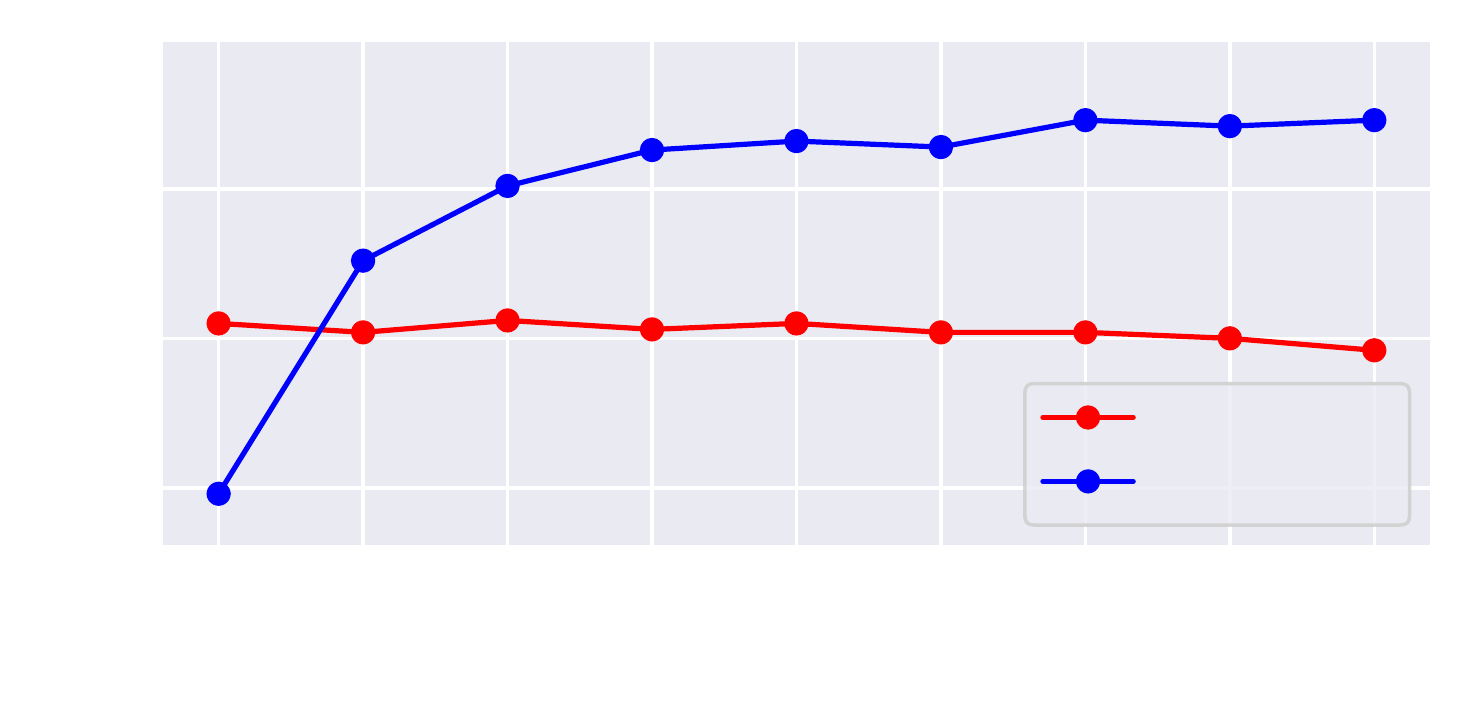} 
    \caption{Impact of varying the ratio of original and synthetic samples used for training the model.}
    \label{fig:dataset_ratio}
\end{figure}

\mypara{Impact of varying the ratio of original and synthetic data.} To determine the best experimental configuration for improving the model's performance with the synthetic dataset $\gD'$, we experiment with different ratios of synthetic data included for training of the CLIP model in the COCO dataset. 
Concretely, we first generate a synthetic dataset using all 5 original captions per image, which is then $|\gD'| / |\gD|. \approx 0.4$. We use the full original dataset $\gD$, but vary the subset from the synthetic dataset $\gD'_{sub} \subset \gD'$ and use both $\gD$ and $\gD'_{sub}$ for training the model. We show the R@1 and ODmAP@1 as we change the amount of synthetic data used for training in Figure \ref{fig:dataset_ratio}. We observe that the ODmAP@1 increases drastically until $|\gD'_{sub}| / |\gD|. \approx 0.1$, and then slowly increases and saturates after the ratio is larger than 0.2. Given that using one out of five original captions per image to generate a synthetic dataset would have $|\gD'_{sub}| / |\gD|. \approx 0.08$ which is near 0.1, we conclude that using synthetic data for a single caption per image is sufficient. We also observe that the R@1 score does not vary much as we change the ratio of the datasets.

\begin{table}[!t]
\centering
\setlength{\tabcolsep}{2.pt}
\resizebox{\columnwidth}{!}{
\renewcommand{\arraystretch}{1.3}
\scriptsize
\begin{tabular}{lcccccccc}
\thickhline
\multirow{2}{*}{Method} & & 
\multirow{2}{*}{\begin{tabular}[c]{@{}c@{}}Text\\ Synth.\end{tabular}} & & 
\multirow{2}{*}{\begin{tabular}[c]{@{}c@{}}Image\\ Synth.\end{tabular}} 
& & \multicolumn{3}{c}{Object decorrelation evaluation}  \\
\\[-2ex]
& & & & & & ODmAP@1 & ODmAP@5 & ODmAP@10  \\ \hline
$\gD$ & & - & & - & & 59.8 & 53.2 & 48.8 \\ \hdashline
\multirow{6}{*}{$\gD + \gD'$} 
&& NP removal & & blur pad. & & 67.9 & 60.9 & 55.8 \\
&& NP removal & & zero pad. & & 68.9 & 61.0 & 55.7 \\ 
&& NP removal & & avg pad. & & 69.1 & 62.0 & 57.1 \\
&& capt. model & & inpaint & & 61.9 & 55.6 & 51.6 \\
&& prompts & & inpaint & & \bf 73.8 & 56.2 & 49.5 \\
\cdashline{3-9}
&& NP removal & & inpaint & & 70.1 & \bf 63.7 & \bf 58.5 \\ \thickhline
\end{tabular}
}
\caption{Comparing the impact of different methods for synthetically generating text and images in $\gD'$ from the MS-COCO dataset on CLIP (finetuned) in terms of its spuriousness.}
\label{tbl:diff_synth}
\vspace{5pt}
\end{table}

\section{Conclusion}
\label{sec:conclusion}

In this paper, we studied the learnt bias in the image-text retrieval models that arises from spurious correlations in the training data. We discovered that existing methods tend to retrieve samples based on cues that might not be semantically related. To address this issue, we trained the model with additional synthetic data which eliminates frequent object co-occurrences in the original training data. Additionally, we proposed a new object decorrelation metric, ODmAP@k, which measures how well the model retrieves samples based on the right cue. Applying our method shows significant improvements on ODmAP@k for a variety of image-text retrieval models, without harming the standard retrieval performance.

\mypara{Limitation and future work.} Our method focuses on data augmentation for the image-text retrieval task. It alleviates the bias caused by spurious correlations between objects in the training data, but does not analyze and solve other biases, e.g.\ texture bias~\cite{geirhos2018imagenet}. Finding and analyzing other biases in the retrieval model would be an interesting topic. Another promising future direction would be to further explore the spuriousness of other cross-modal retrieval tasks such as text-video retrieval or text-audio retrieval. 

\section*{Acknowledgements}
This work was supported by DFG project number 276693517, by BMBF FKZ: 01IS18039A, by the ERC (853489 - DEXIM), and by EXC number 2064/1 – project number 390727645.
Jae Myung Kim thanks the European Laboratory for Learning and Intelligent Systems (ELLIS) PhD program and the International Max
Planck Research School for Intelligent Systems (IMPRS-IS) for support.

\clearpage
{\small
\bibliographystyle{ieee_fullname}
\bibliography{main}

\begin{thebibliography}{10}\itemsep=-1pt

\bibitem{agarwal2020towards}
Vedika Agarwal, Rakshith Shetty, and Mario Fritz.
\newblock Towards causal vqa: Revealing and reducing spurious correlations by
  invariant and covariant semantic editing.
\newblock In {\em CVPR}, 2020.

\bibitem{alayrac2022flamingo}
Jean-Baptiste Alayrac, Jeff Donahue, Pauline Luc, Antoine Miech, Iain Barr,
  Yana Hasson, Karel Lenc, Arthur Mensch, Katie Millican, Malcolm Reynolds,
  et~al.
\newblock Flamingo: a visual language model for few-shot learning.
\newblock {\em arXiv preprint arXiv:2204.14198}, 2022.

\bibitem{aytar2008utilizing}
Yusuf Aytar, Mubarak Shah, and Jiebo Luo.
\newblock Utilizing semantic word similarity measures for video retrieval.
\newblock In {\em CVPR}, 2008.

\bibitem{bain2021frozen}
Max Bain, Arsha Nagrani, G{\"u}l Varol, and Andrew Zisserman.
\newblock Frozen in time: A joint video and image encoder for end-to-end
  retrieval.
\newblock In {\em ICCV}, 2021.

\bibitem{berg2022prompt}
Hugo Berg, Siobhan~Mackenzie Hall, Yash Bhalgat, Wonsuk Yang, Hannah~Rose Kirk,
  Aleksandar Shtedritski, and Max Bain.
\newblock A prompt array keeps the bias away: Debiasing vision-language models
  with adversarial learning.
\newblock {\em arXiv preprint arXiv:2203.11933}, 2022.

\bibitem{bogolin2022cross}
Simion-Vlad Bogolin, Ioana Croitoru, Hailin Jin, Yang Liu, and Samuel Albanie.
\newblock Cross modal retrieval with querybank normalisation.
\newblock In {\em CVPR}, 2022.

\bibitem{chang2018explaining}
Chun-Hao Chang, Elliot Creager, Anna Goldenberg, and David Duvenaud.
\newblock Explaining image classifiers by counterfactual generation.
\newblock In {\em ICLR}, 2018.

\bibitem{chen2020counterfactual}
Long Chen, Xin Yan, Jun Xiao, Hanwang Zhang, Shiliang Pu, and Yueting Zhuang.
\newblock Counterfactual samples synthesizing for robust visual question
  answering.
\newblock In {\em CVPR}, 2020.

\bibitem{chen2020adaptive}
Tianlang Chen, Jiajun Deng, and Jiebo Luo.
\newblock Adaptive offline quintuplet loss for image-text matching.
\newblock In {\em ECCV}, 2020.

\bibitem{cho2014properties}
Kyunghyun Cho, Bart Van~Merri{\"e}nboer, Dzmitry Bahdanau, and Yoshua Bengio.
\newblock On the properties of neural machine translation: Encoder-decoder
  approaches.
\newblock {\em arXiv preprint arXiv:1409.1259}, 2014.

\bibitem{chun2022eccv_caption}
Sanghyuk Chun, Wonjae Kim, Song Park, Minsuk~Chang Chang, and Seong~Joon Oh.
\newblock Eccv caption: Correcting false negatives by collecting
  machine-and-human-verified image-caption associations for ms-coco.
\newblock In {\em ECCV}, 2022.

\bibitem{chun2021probabilistic}
Sanghyuk Chun, Seong~Joon Oh, Rafael~Sampaio De~Rezende, Yannis Kalantidis, and
  Diane Larlus.
\newblock Probabilistic embeddings for cross-modal retrieval.
\newblock In {\em ICCV}, 2021.

\bibitem{croitoru2021teachtext}
Ioana Croitoru, Simion-Vlad Bogolin, Marius Leordeanu, Hailin Jin, Andrew
  Zisserman, Samuel Albanie, and Yang Liu.
\newblock Teachtext: Crossmodal generalized distillation for text-video
  retrieval.
\newblock In {\em ICCV}, 2021.

\bibitem{crowson2022vqgan}
Katherine Crowson, Stella Biderman, Daniel Kornis, Dashiell Stander, Eric
  Hallahan, Louis Castricato, and Edward Raff.
\newblock Vqgan-clip: Open domain image generation and editing with natural
  language guidance.
\newblock In {\em ECCV}, 2022.

\bibitem{deng2009imagenet}
Jia Deng, Wei Dong, Richard Socher, Li-Jia Li, Kai Li, and Li Fei-Fei.
\newblock Imagenet: A large-scale hierarchical image database.
\newblock In {\em CVPR}, 2009.

\bibitem{devlin2018bert}
Jacob Devlin, Ming-Wei Chang, Kenton Lee, and Kristina Toutanova.
\newblock Bert: Pre-training of deep bidirectional transformers for language
  understanding.
\newblock 2018.

\bibitem{dong2018predicting}
Jianfeng Dong, Xirong Li, and Cees~GM Snoek.
\newblock Predicting visual features from text for image and video caption
  retrieval.
\newblock {\em IEEE Transactions on Multimedia}, 2018.

\bibitem{dosovitskiy2020image}
Alexey Dosovitskiy, Lucas Beyer, Alexander Kolesnikov, Dirk Weissenborn,
  Xiaohua Zhai, Thomas Unterthiner, Mostafa Dehghani, Matthias Minderer, Georg
  Heigold, Sylvain Gelly, et~al.
\newblock An image is worth 16x16 words: Transformers for image recognition at
  scale.
\newblock In {\em iclr}, 2020.

\bibitem{eisenschtat2017linking}
Aviv Eisenschtat and Lior Wolf.
\newblock Linking image and text with 2-way nets.
\newblock In {\em CVPR}, 2017.

\bibitem{engilberge2018finding}
Martin Engilberge, Louis Chevallier, Patrick P{\'e}rez, and Matthieu Cord.
\newblock Finding beans in burgers: Deep semantic-visual embedding with
  localization.
\newblock In {\em CVPR}, 2018.

\bibitem{faghri2017vse}
Fartash Faghri, David~J Fleet, Jamie~Ryan Kiros, and Sanja Fidler.
\newblock Vse++: Improving visual-semantic embeddings with hard negatives.
\newblock In {\em BMVC}, 2018.

\bibitem{frome2013devise}
Andrea Frome, Greg~S Corrado, Jon Shlens, Samy Bengio, Jeff Dean, Marc'Aurelio
  Ranzato, and Tomas Mikolov.
\newblock Devise: A deep visual-semantic embedding model.
\newblock In {\em NeurIPS}, 2013.

\bibitem{gabeur2022masking}
Valentin Gabeur, Arsha Nagrani, Chen Sun, Karteek Alahari, and Cordelia Schmid.
\newblock Masking modalities for cross-modal video retrieval.
\newblock 2022.

\bibitem{gabeur2020multi}
Valentin Gabeur, Chen Sun, Karteek Alahari, and Cordelia Schmid.
\newblock Multi-modal transformer for video retrieval.
\newblock In {\em ECCV}, 2020.

\bibitem{gao2022pyramidclip}
Yuting Gao, Jinfeng Liu, Zihan Xu, Jun Zhang, Ke Li, and Chunhua Shen.
\newblock Pyramidclip: Hierarchical feature alignment for vision-language model
  pretraining.
\newblock {\em arXiv preprint arXiv:2204.14095}, 2022.

\bibitem{geirhos2018imagenet}
Robert Geirhos, Patricia Rubisch, Claudio Michaelis, Matthias Bethge, Felix~A
  Wichmann, and Wieland Brendel.
\newblock Imagenet-trained cnns are biased towards texture; increasing shape
  bias improves accuracy and robustness.
\newblock In {\em ICLR}, 2019.

\bibitem{gong2014improving}
Yunchao Gong, Liwei Wang, Micah Hodosh, Julia Hockenmaier, and Svetlana
  Lazebnik.
\newblock Improving image-sentence embeddings using large weakly annotated
  photo collections.
\newblock In {\em ECCV}, 2014.

\bibitem{goyal2017making}
Yash Goyal, Tejas Khot, Douglas Summers-Stay, Dhruv Batra, and Devi Parikh.
\newblock Making the v in vqa matter: Elevating the role of image understanding
  in visual question answering.
\newblock In {\em CVPR}, 2017.

\bibitem{harwath2018jointly}
David Harwath, Adria Recasens, D{\'\i}dac Sur{\'\i}s, Galen Chuang, Antonio
  Torralba, and James Glass.
\newblock Jointly discovering visual objects and spoken words from raw sensory
  input.
\newblock In {\em ECCV}, 2018.

\bibitem{hendricks2018women}
Lisa~Anne Hendricks, Kaylee Burns, Kate Saenko, Trevor Darrell, and Anna
  Rohrbach.
\newblock Women also snowboard: Overcoming bias in captioning models.
\newblock In {\em ECCV}, 2018.

\bibitem{hong2017deep}
Sungeun Hong, Woobin Im, and Hyun~S Yang.
\newblock Deep learning for content-based, cross-modal retrieval of videos and
  music.
\newblock {\em arXiv preprint arXiv:1704.06761}, 2017.

\bibitem{jia2021scaling}
Chao Jia, Yinfei Yang, Ye Xia, Yi-Ting Chen, Zarana Parekh, Hieu Pham, Quoc Le,
  Yun-Hsuan Sung, Zhen Li, and Tom Duerig.
\newblock Scaling up visual and vision-language representation learning with
  noisy text supervision.
\newblock 2021.

\bibitem{karpathy2015deep}
Andrej Karpathy and Li Fei-Fei.
\newblock Deep visual-semantic alignments for generating image descriptions.
\newblock In {\em CVPR}, 2015.

\bibitem{karpathy2014deep}
Andrej Karpathy, Armand Joulin, and Li~F Fei-Fei.
\newblock Deep fragment embeddings for bidirectional image sentence mapping.
\newblock 2014.

\bibitem{kingma2014adam}
Diederik~P Kingma and Jimmy Ba.
\newblock Adam: A method for stochastic optimization.
\newblock {\em arXiv preprint arXiv:1412.6980}, 2014.

\bibitem{kiros2014unifying}
Ryan Kiros, Ruslan Salakhutdinov, and Richard~S Zemel.
\newblock Unifying visual-semantic embeddings with multimodal neural language
  models.
\newblock {\em arXiv preprint arXiv:1411.2539}, 2014.

\bibitem{koepke2022audio}
A~Sophia Koepke, Andreea-Maria Oncescu, Joao Henriques, Zeynep Akata, and
  Samuel Albanie.
\newblock Audio retrieval with natural language queries: A benchmark study.
\newblock {\em IEEE Transactions on Multimedia}, 2022.

\bibitem{lee2018stacked}
Kuang-Huei Lee, Xi Chen, Gang Hua, Houdong Hu, and Xiaodong He.
\newblock Stacked cross attention for image-text matching.
\newblock In {\em ECCV}, 2018.

\bibitem{li2022blip}
Junnan Li, Dongxu Li, Caiming Xiong, and Steven Hoi.
\newblock Blip: Bootstrapping language-image pre-training for unified
  vision-language understanding and generation.
\newblock In {\em ICML}, 2022.

\bibitem{li2019visual}
Kunpeng Li, Yulun Zhang, Kai Li, Yuanyuan Li, and Yun Fu.
\newblock Visual semantic reasoning for image-text matching.
\newblock In {\em ICCV}, 2019.

\bibitem{li2020oscar}
Xiujun Li, Xi Yin, Chunyuan Li, Pengchuan Zhang, Xiaowei Hu, Lei Zhang, Lijuan
  Wang, Houdong Hu, Li Dong, Furu Wei, et~al.
\newblock Oscar: Object-semantics aligned pre-training for vision-language
  tasks.
\newblock In {\em ECCV}, 2020.

\bibitem{lin2014microsoft}
Tsung-Yi Lin, Michael Maire, Serge Belongie, James Hays, Pietro Perona, Deva
  Ramanan, Piotr Doll{\'a}r, and C~Lawrence Zitnick.
\newblock Microsoft coco: Common objects in context.
\newblock In {\em ECCV}, 2014.

\bibitem{loshchilov2018decoupled}
Ilya Loshchilov and Frank Hutter.
\newblock Decoupled weight decay regularization.
\newblock In {\em ICLR}, 2018.

\bibitem{lou2022audio}
Siyu Lou, Xuenan Xu, Mengyue Wu, and Kai Yu.
\newblock Audio-text retrieval in context.
\newblock In {\em ICASSP}, 2022.

\bibitem{lu2019vilbert}
Jiasen Lu, Dhruv Batra, Devi Parikh, and Stefan Lee.
\newblock Vilbert: Pretraining task-agnostic visiolinguistic representations
  for vision-and-language tasks.
\newblock {\em NeurIPS}, 2019.

\bibitem{luo2022clip4clip}
Huaishao Luo, Lei Ji, Ming Zhong, Yang Chen, Wen Lei, Nan Duan, and Tianrui Li.
\newblock Clip4clip: An empirical study of clip for end to end video clip
  retrieval and captioning.
\newblock {\em Neurocomputing}, 2022.

\bibitem{ma2022ei}
Haoyu Ma, Handong Zhao, Zhe Lin, Ajinkya Kale, Zhangyang Wang, Tong Yu,
  Jiuxiang Gu, Sunav Choudhary, and Xiaohui Xie.
\newblock Ei-clip: Entity-aware interventional contrastive learning for
  e-commerce cross-modal retrieval.
\newblock In {\em CVPR}, 2022.

\bibitem{DCLIP}
Sachit Menon and Carl Vondrick.
\newblock Visual classification via description from large language models.
\newblock In {\em ICLR}, 2023.

\bibitem{mithun2018learning}
Niluthpol~Chowdhury Mithun, Juncheng Li, Florian Metze, and Amit~K
  Roy-Chowdhury.
\newblock Learning joint embedding with multimodal cues for cross-modal
  video-text retrieval.
\newblock In {\em ACM International Conference on Multimedia Retrieval}, 2018.

\bibitem{mokady2021clipcap}
Ron Mokady, Amir Hertz, and Amit~H Bermano.
\newblock Clipcap: Clip prefix for image captioning.
\newblock {\em arXiv preprint arXiv:2111.09734}, 2021.

\bibitem{mu2022slip}
Norman Mu, Alexander Kirillov, David Wagner, and Saining Xie.
\newblock Slip: Self-supervision meets language-image pre-training.
\newblock In {\em ECCV}, 2022.

\bibitem{musgrave2020metric}
Kevin Musgrave, Serge Belongie, and Ser-Nam Lim.
\newblock A metric learning reality check.
\newblock In {\em ECCV}, 2020.

\bibitem{nagrani2018learnable}
Arsha Nagrani, Samuel Albanie, and Andrew Zisserman.
\newblock Learnable pins: Cross-modal embeddings for person identity.
\newblock In {\em ECCV}, 2018.

\bibitem{nagrani2022learning}
Arsha Nagrani, Paul~Hongsuck Seo, Bryan Seybold, Anja Hauth, Santiago Manen,
  Chen Sun, and Cordelia Schmid.
\newblock Learning audio-video modalities from image captions.
\newblock {\em arXiv preprint arXiv:2204.00679}, 2022.

\bibitem{niu2021counterfactual}
Yulei Niu, Kaihua Tang, Hanwang Zhang, Zhiwu Lu, Xian-Sheng Hua, and Ji-Rong
  Wen.
\newblock Counterfactual vqa: A cause-effect look at language bias.
\newblock In {\em CVPR}, 2021.

\bibitem{oncescu2021audio}
Andreea-Maria Oncescu, A Koepke, Joao~F Henriques, Zeynep Akata, and Samuel
  Albanie.
\newblock Audio retrieval with natural language queries.
\newblock In {\em INTERSPEECH}, 2021.

\bibitem{CUPL}
Sarah Pratt, Rosanne Liu, and Ali Farhadi.
\newblock What does a platypus look like? generating customized prompts for
  zero-shot image classification.
\newblock {\em arXiv preprint arXiv:2209.03320}, 2022.

\bibitem{radford2021learning}
Alec Radford, Jong~Wook Kim, Chris Hallacy, Aditya Ramesh, Gabriel Goh,
  Sandhini Agarwal, Girish Sastry, Amanda Askell, Pamela Mishkin, Jack Clark,
  et~al.
\newblock Learning transferable visual models from natural language
  supervision.
\newblock 2021.

\bibitem{radford2019language}
Alec Radford, Jeffrey Wu, Rewon Child, David Luan, Dario Amodei, Ilya
  Sutskever, et~al.
\newblock Language models are unsupervised multitask learners.
\newblock {\em OpenAI blog}, 1(8):9, 2019.

\bibitem{rohrbach2018object}
Anna Rohrbach, Lisa~Anne Hendricks, Kaylee Burns, Trevor Darrell, and Kate
  Saenko.
\newblock Object hallucination in image captioning.
\newblock In {\em EMNLP}, 2018.

\bibitem{shin2022reco}
Gyungin Shin, Weidi Xie, and Samuel Albanie.
\newblock Reco: Retrieve and co-segment for zero-shot transfer.
\newblock In {\em NeurIPS}, 2022.

\bibitem{simonyan2014very}
Karen Simonyan and Andrew Zisserman.
\newblock Very deep convolutional networks for large-scale image recognition.
\newblock {\em arXiv preprint arXiv:1409.1556}, 2014.

\bibitem{song2019polysemous}
Yale Song and Mohammad Soleymani.
\newblock Polysemous visual-semantic embedding for cross-modal retrieval.
\newblock In {\em CVPR}, 2019.

\bibitem{thomas2020preserving}
Christopher Thomas and Adriana Kovashka.
\newblock Preserving semantic neighborhoods for robust cross-modal retrieval.
\newblock In {\em ECCV}, 2020.

\bibitem{wang2021gender}
Jialu Wang, Yang Liu, and Xin Wang.
\newblock Are gender-neutral queries really gender-neutral? mitigating gender
  bias in image search.
\newblock In {\em EMNLP}, 2021.

\bibitem{wang2016learning}
Liwei Wang, Yin Li, and Svetlana Lazebnik.
\newblock Learning deep structure-preserving image-text embeddings.
\newblock In {\em CVPR}, 2016.

\bibitem{wang2019camp}
Zihao Wang, Xihui Liu, Hongsheng Li, Lu Sheng, Junjie Yan, Xiaogang Wang, and
  Jing Shao.
\newblock Camp: Cross-modal adaptive message passing for text-image retrieval.
\newblock In {\em ICCV}, 2019.

\bibitem{wray2019fine}
Michael Wray, Diane Larlus, Gabriela Csurka, and Dima Damen.
\newblock Fine-grained action retrieval through multiple parts-of-speech
  embeddings.
\newblock In {\em ICCV}, 2019.

\bibitem{xu2015jointly}
Ran Xu, Caiming Xiong, Wei Chen, and Jason Corso.
\newblock Jointly modeling deep video and compositional text to bridge vision
  and language in a unified framework.
\newblock In {\em AAAI}, 2015.

\bibitem{yao2021filip}
Lewei Yao, Runhui Huang, Lu Hou, Guansong Lu, Minzhe Niu, Hang Xu, Xiaodan
  Liang, Zhenguo Li, Xin Jiang, and Chunjing Xu.
\newblock Filip: Fine-grained interactive language-image pre-training.
\newblock {\em arXiv preprint arXiv:2111.07783}, 2021.

\bibitem{young2014image}
Peter Young, Alice Lai, Micah Hodosh, and Julia Hockenmaier.
\newblock From image descriptions to visual denotations: New similarity metrics
  for semantic inference over event descriptions.
\newblock {\em Transactions of the Association for Computational Linguistics},
  2014.

\bibitem{yu2018generative}
Jiahui Yu, Zhe Lin, Jimei Yang, Xiaohui Shen, Xin Lu, and Thomas~S Huang.
\newblock Generative image inpainting with contextual attention.
\newblock In {\em CVPR}, 2018.

\bibitem{zhang2020context}
Qi Zhang, Zhen Lei, Zhaoxiang Zhang, and Stan~Z Li.
\newblock Context-aware attention network for image-text retrieval.
\newblock In {\em CVPR}, 2020.

\bibitem{zhang2020devlbert}
Shengyu Zhang, Tan Jiang, Tan Wang, Kun Kuang, Zhou Zhao, Jianke Zhu, Jin Yu,
  Hongxia Yang, and Fei Wu.
\newblock Devlbert: Learning deconfounded visio-linguistic representations.
\newblock In {\em ACM MM}, 2020.

\bibitem{zhao2021understanding}
Dora Zhao, Angelina Wang, and Olga Russakovsky.
\newblock Understanding and evaluating racial biases in image captioning.
\newblock In {\em ICCV}, 2021.

\end{thebibliography}
}

\clearpage

\section*{A. Matching table between noun phrases and class names.} 
When synthesizing the text by removing noun phrase chunks, we should match the noun phrase with the class names of the object to be removed. While this (class name, noun phrase) pair is annotated in the Flickr30k dataset, we manually list the matching pair in the MS-COCO dataset. If the noun phrase contains a word related to the given class name, we regard that noun phrase as matching the given class name. The matching table is given in Table \ref{tbl:appendix:match}. These matching pairs are based on the implementation done in the  previous literature \cite{agarwal2020towards}, but we added and removed some pairs to make the pairs more relevant. For brevity, we did not list the word that is identical to the class name on the right-hand side of the table. 

\section*{B. Analysis of distribution shift between the synthetic ($D’$) and the original ($D$) datasets.}
\begin{table}[H]
\centering\resizebox{0.3\textwidth}{!}{
\vspace{-5pt}
\begin{tabular}{lccc}
        CLIP && ODmAP@1 & i2t R@1 \\
        \cline{1-1}\cline{3-4}
        \vspace{-1em} & \\
        \cline{1-1}\cline{3-4}
        \vspace{-1em} & \\
        zero-shot && 58.6 & 50.6  \\
        $D_s$ && 61.5 & 60.5 \\
        $D'$ && 66.4 & 58.1  \\
        $D\!\!+\!\!D'$ && 70.1 & 65.6  \\
        \cline{1-1}\cline{3-4}
    \end{tabular}
}
\vspace{-5pt}
\end{table}
As $|D’| < |D|$ (one-third smaller), we made a new dataset $D_s \subset D$ where $|D_s| = |D’|$ for comparison. Finetuning CLIP with $D’$ and $D_s$, respectively, resulted in pretty similar results (differing by 2.4\% i2t R@1). Considering the 9.9\% improvement from zero-shot to $D_s$, the data distribution of $D’$ seems not much shifted from the data distribution of $D$ even with somewhat broken visual and linguistic coherence in $D’$. 
Also, compared to $D\!+\!D'$, $D'$ lowers ODmAP@1 by 3.7\%. We think this is because information on de-correlated objects in $D$ is not learned by the model trained only with $D'$.   

\section*{C. Pseudo-code.} We present the pseudo-code for the implementation of our proposed data synthesis in Listing \ref{pseudocode:appendix}.

{
\setlength{\tabcolsep}{7pt}
\renewcommand{\arraystretch}{1.2}
\begin{table*}[t]
\footnotesize
\centering
\begin{tabular}{ll}
\hline
\begin{tabular}[l]{@{}l@{}}Class name \\in MS-COCO\end{tabular} & Word in noun phrase chunk \\ \hline
person & \begin{tabular}[l]{@{}l@{}} 
man, woman, player, child, girl, boy, boys, people, lady, guy, kid, kids, surfer, cowboy, cowboys, \\
adult, adults, cop, soldier, police, catcher, pitcher, jockey, baby, men, women, biker, spectator, rider, \\
batter, gay, anyone, someone, reporter, somebody, anybody, everyone, worker, workers \end{tabular} \\
airplane & plane, jet, aircraft \\
bicycle & bike, biking, cycling \\
motorcycle & motor \\
bus & trolley \\
car & van, taxi, trunk, truck, suv \\
train & tram, subway \\
traffic light & traffic \\
stop sign & sign \\
parking meter & meter \\
fire hydrant & hydrant, hydrate, hydra \\
bird & beak, duck, goose, gull, pigeon, chicken, penguin \\
cat & kitty, kitten \\
dog & puppy, puppies \\
sheep & lamb \\
horse & pony, foal \\
cow & cattle, oxen, ox, herd, calves, bull, calf \\
handbag & bag \\
suitcase & bag, luggage, case \\
frisbee & disc, disk, frisby \\
sports ball & ball \\
baseball bat & bat \\
baseball glove & glove \\
skateboard & board, skate \\
surfboard & board \\
snowboard & board \\
skis & ski \\
tennis racket & racket, racquet \\
wine glass & glass, wine, beverage \\
bottle & thermos, flask, beer, beverage \\
cup & glass, mug, beverage, coffee, tea \\
spoon & siverware \\
donut & doughnut, dough \\
cake & dessert, frosting \\
dining table & desk, table, tables \\
chair & stool \\
potted plant & plant, flower \\
vase & pot, vase \\
tv & television, screen \\
laptop & computer, monitor, screen \\
cell phone & phone \\
refrigerator & fridge \\
book & novel \\
scissors & scissor \\
toothbrush & brush \\
hair drier & drier \\
teddy bear & teddy, toy, bear, doll \\
\hline
\end{tabular}
\caption{\textbf{Matching table between class names and noun phrases.} We regard the noun phrase as matching the given class name if the word related to the class name is contained in the noun phrase.}
\vspace{-5pt}
\label{tbl:appendix:match}
\end{table*}
}

\clearpage
\noindent\begin{minipage}{\textwidth}
\begin{lstlisting}[language=Python, caption=Pseudo-code for the proposed data synthesis method to reduce spuriousness., label={pseudocode:appendix}]
# Threshold for data synthesis (Section 3.1)
alpha1 = 0.4
alpha2 = 0.8
alpha3 = 0.7

# get synthetic data 
for image_idx in ranger(n_images):
    image, caption, bboxes, bbox_classnames = dataset.__getitem__(image_idx)
    
    # extract nounphrases from caption using NLTK tool
    nounphrases = get_nounphrases(caption)

    # get masks for each classname in the image
    classname_set = list(set(bbox_classnames))
    mask_list = []
    for classname_to_remove in classname_set:
        bbox_idxs_to_remove = [i for i, _cat in enumerate(bbox_classnames) 
                               if _cat == classname_to_remove]
        bboxes_to_remove = bboxes[bbox_idxs_to_remove]
        mask = union_bboxes(bboxes_to_remove) # union all the bboxes
        mask_list.append(mask)

    if len(classnames_set) >= 2:
        for i, classname_to_remove in enumerate(classname_set):
            # classname_to_remove to be removed
            mask_q = mask_list[i]
            mask_gs = [mask for j, mask in enumerate(mask_list) if j != i]
            classname_gs = [_c for j, _c in enumerate(classname_set) if j != i]

            # check size of removed region
            if mask_q.sum() / (mask_q.size(2) * mask_q.size(3)) > alpha3:
                continue
                    
            # check overlap between bbox from selected class and others
            overlaps = torch.tensor(
                [torch.logical_and(mask_q, mask_g).sum() / mask_g.sum() 
                 for mask_g in mask_gs])

            # when removing a single class
            if all(overlaps < alpha1):
                # synthetic image using inpainting GAN
                synth_image = synthesize_image(image, mask_q) 
                # synthetic caption using matching table in Appendix
                synth_caption = synthesize_caption(caption, classname_to_remove)
                
            # when removing multiple classes
            elif any(overlaps > alpha2):
                bool_overlaps = overlaps > alpha2
                mask_qs = \
                    [mask_q] + [_m for j, _m in enumerate(mask_gs) if bool_overlaps[j]]
                classnames_to_remove = \
                    [classname] + [_c for j, _c in enumerate(classname_gs) if bool_overlaps[j]]
                # synthetic image using inpainting GAN
                synth_image = synthesize_image(image, mask_qs)
                # synthetic caption using matching table in Appendix
                synth_caption = synthesize_caption(caption, classnames_to_remove)
\end{lstlisting}
\end{minipage}

\end{document}